\documentclass[sigconf]{acmart}
\usepackage[utf8]{inputenc}
\AtBeginDocument{%
  \providecommand\BibTeX{{%
    \normalfont B\kern-0.5em{\scshape i\kern-0.25em b}\kern-0.8em\TeX}}}

\setcopyright{acmcopyright}
\copyrightyear{2018}
\acmYear{2018}
\acmDOI{10.1145/1122445.1122456}

\acmConference[Gold Coast '21]{Gold Coast '21: ACM Multimedia Asia}{Dec 01--03, 2021}{Gold Coast, Australia}
\acmBooktitle{Gold Coast '21: ACM Multimedia Asia, Dec 01--03, 2018, Gold Coast, Australia}
\acmPrice{15.00}
\acmISBN{978-1-4503-XXXX-X/18/06}

\begin{document}

\title{RoadAtlas: Intelligent Platform for Automated Road Defect Detection and Asset Management}

\author{
  Zhuoxiao Chen$^1$,
  Yiyun Zhang$^1$,
  Yadan Luo$^1$,
  Zijian Wang$^1$, 
  Jinjiang Zhong$^2$,
  Anthony Southon$^2$
}
\affiliation{
  \institution{$^1$The University of Queensland, $^2$Logan City Council, Australia}
  \city{}
  \country{}}
\email{{zhuoxiao.chen, yiyun.zhang, y.luo, zijian.wang}@uq.edu.au, 
{JinjiangZhong, AnthonySouthon}@logan.qld.gov.au}


\begin{abstract}
With the rapid development of intelligent detection algorithms based on deep learning, much progress has been made in automatic road defect recognition and road marking parsing. This can effectively address the issue of an expensive and time-consuming process for professional inspectors to review the street manually. Towards this goal, we present RoadAtlas, a novel end-to-end integrated system that can support 1) road defect detection, 2) road marking parsing, 3) a web-based dashboard for presenting and inputting data by users, and 4) a backend containing a well-structured database and developed APIs.

\end{abstract}




\maketitle


\section{Introduction}

Assessing the condition of the road network is vital for ensuring its continued functionality. The majority of road agencies in Australia conduct road network condition surveys through employing specialised machines such as laser profiler meters to obtain performance and condition data of the road network. There are also a number of road agencies that rely on experience inspectors via walking or driving the road network to collect defects data. Being a costly and time-consuming procedure (around \$350K to \$400K per survey), it prevents the local government from conducting frequent surveys to keep the data up to date. The cracks and destroyed road markings not only affect the aesthetic value of the city, but also increase the risks of injury and fatality faced by pedestrians. Hence, there is an urgent need for automation in road asset management by adopting data science for predictive analytics, visual recognition, and interactive visualisation. 

The framework’s purpose is to forge a mutually beneficial program by forming a novel AI-powered road defects and assets management platform, namely \textbf{RoadAtlas}. Our vision is to deliver the intelligent system for jointly road defect segmentation and road marking parsing, being able to grow and adapt to the new data or novel tasks (e.g., detecting a new crack type). The RoadAtlas will afford an unprecedented capacity to automatically demonstrate and report types, severity, location and evidence of road anomalies in real-time for planning the follow-up repairs. The RoadAtlas also enhances local government employees’ data analytic skillsets and help road asset managers monitor the functional and structural performance of road networks efficiently and effectively.

\section{System Overview}

\begin{figure}[t]
    \centering
    \includegraphics[width=\linewidth]{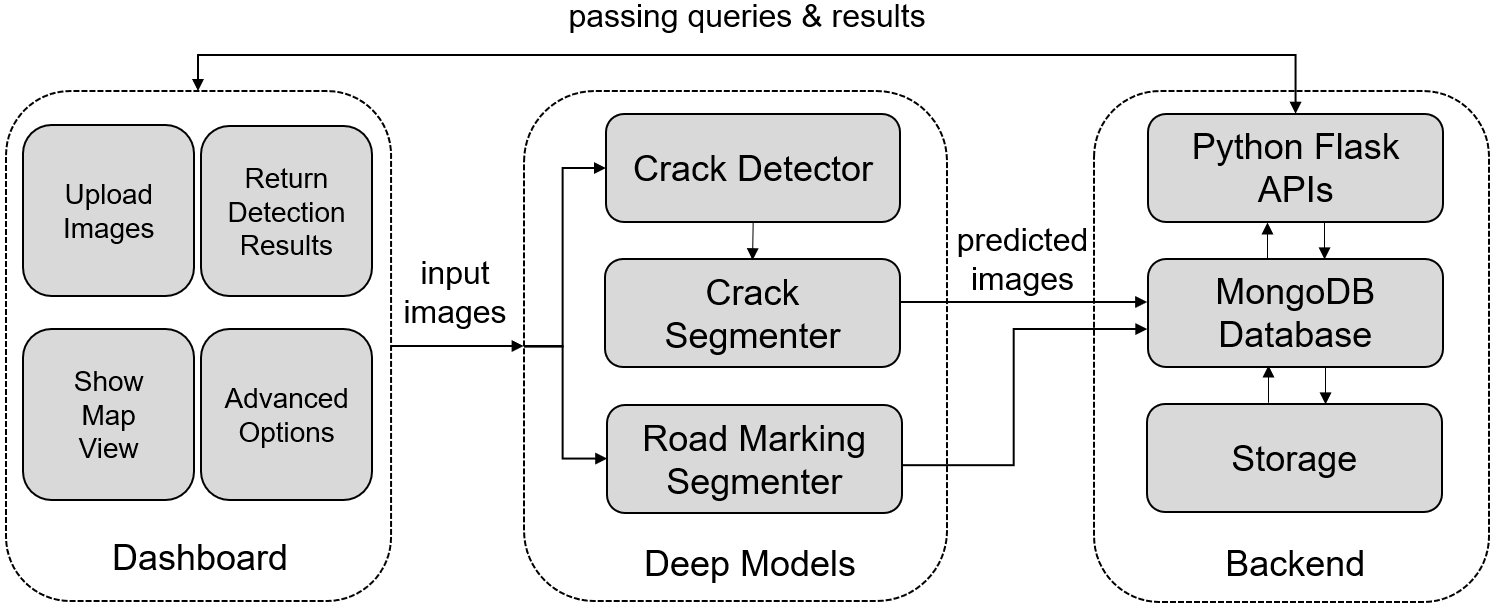}\vspace{-2ex}
    \caption{The overall framework of the RoadAtlas. \vspace{-2ex}}
    \label{fig:framework}
\end{figure}

The system consists of three modules: the dashboard, deep visual models and the backend, as shown in Figure \ref{fig:framework}. The dashboard is a user interaction interface that supports various front-end functions. The deep models are designed for road defect and road marking recognition. The backend is built with Python Flask to exchange data with the front-end and save data in the MongoDB database.
\vspace{-2ex}
\begin{figure}[!ht]
    \centering
    \includegraphics[width=\linewidth]{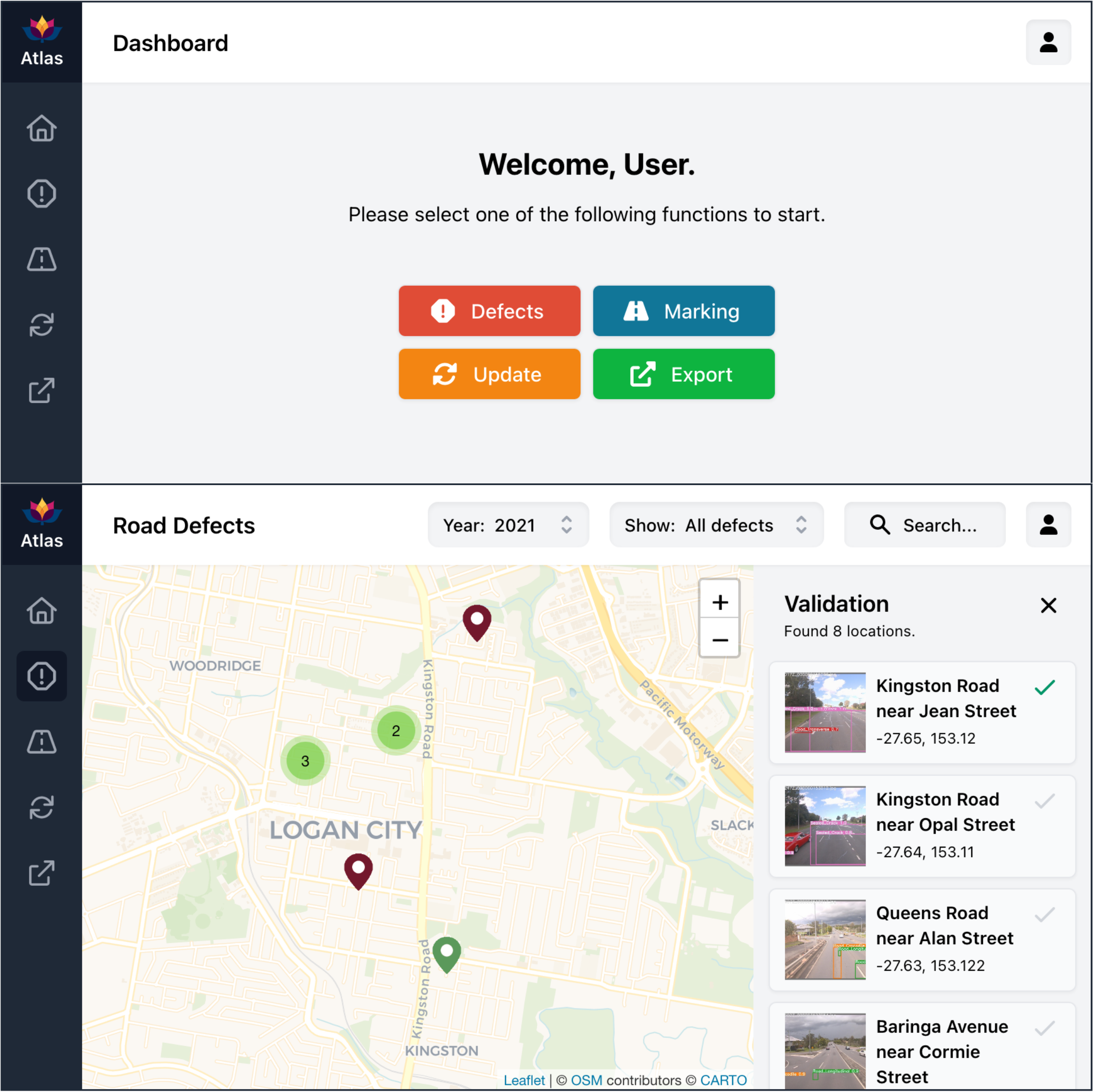}\vspace{-1ex}
    \caption{Dashboard (top) and Road Defects View (bottom).\vspace{-4ex}}
    \label{fig:dashboard}
\end{figure}
\subsection{Front-end and Back-end}
The Dashboard is the primary user interface designed as a web application, illustrated in Figure \ref{fig:dashboard}. It features four primary functions: 1) {\itshape Road Defect}, 2) {\itshape Road Marking}, 3) {\itshape Database Update} and 4) {\itshape Report Export}. {\itshape Road Defect} and {\itshape Road Marking} allow users to inspect and validate labelled images from the database. The {\itshape Road Defect} example shown in Figure \ref{fig:dashboard} displays an interactive map with map pins representing individual defects. When the user selects a map pin, a validation panel displays on the right, providing defect details such as the labelled images and coordinates. Next, the user can validate the correctness of defects and mark them as checked. The {\itshape Database Update} function is a gateway for users to upload unprocessed images to the deep model for defect and marking detection. Additional {\itshape Report Export} function supports saving validated images data as CSV or JSON-formatted files.
The backend aims to provide functional APIs to exchange data with the dashboard side. Our developers set up multiple APIs according to the need of the dashboard via the Python Flask framework. Also, we construct the MongoDB database for managing data. We further provide 20TB hard disk storage on our server for storing huge amounts of images.



\subsection{Road Defect Detection and Segmentation}
When users utilise the \textit{database update} function by inserting new road images, the proposed deep models can perform recognition for road defects.

\noindent\textbf{Dataset Annotation.}
To train models for road defect detection following supervised learning principles, labels are required for training images. However, the 10,000 raw images provided by the local government are unlabelled. These images represent the foreground of the car, taken in Logan City, Australia. We arranged five annotators to label all images with various classes of road defects, including \textit{background}, \textit{Kerb\_Cracking}, \textit{Road\_Crocodile}, \textit{Road\_Longitudinal}, \textit{Road\_Transverse}, \textit{Road\_Block} and \textit{Sealed\_Crack}. Each of the road defects is labelled in an irregular polygon.




\begin{figure}[th]
    \centering
    \includegraphics[width=0.9\linewidth]{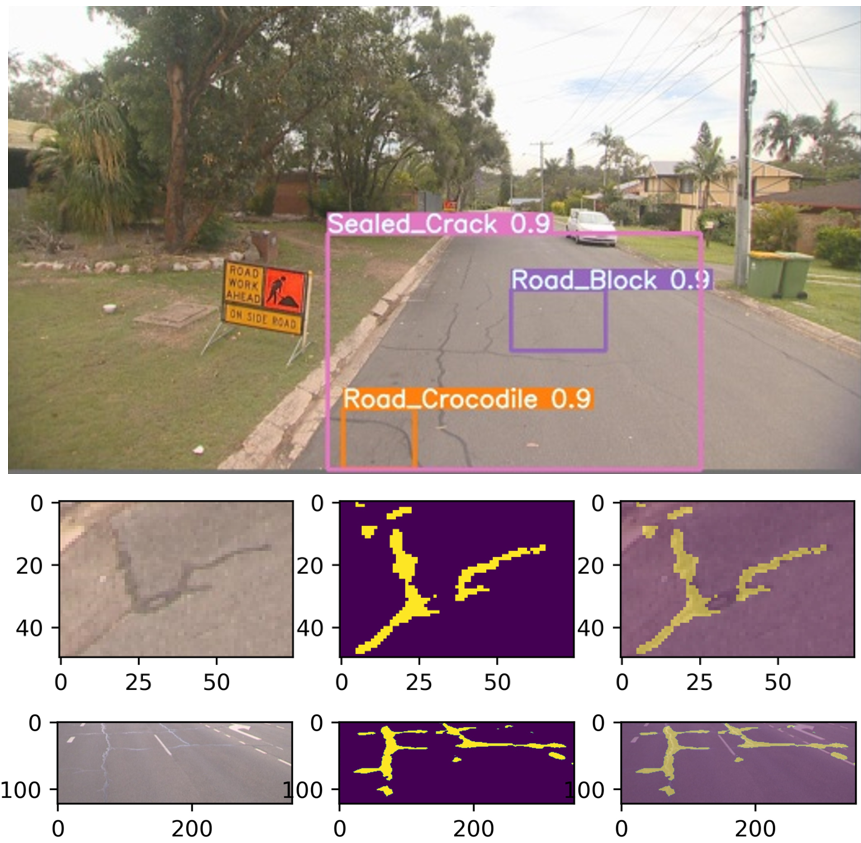}
    \caption{Examples of road defect detection and segmentation results.\vspace{-4ex}}
    \label{fig:seg_result}
\end{figure}
\noindent\textbf{Crack Detection and Segmentation.}
We developed a pair of deep models to effectively capture road defects consisting of a crack detector and a crack segmenter. There are four steps for crack recognition. First, we utilise the anonymiser \cite{Anonymizer} to protect the privacy of car owners via automatically hiding the license plate numbers. Second, we convert labelled polygons to the minimised bounding boxes that cover the polygon. Third, we proposed a crack detector inspired by \cite{glenn_jocher_2020_4154370}, fully trained with converted labelled bounding boxes. Thus, the crack detector can accurately produce bounding boxes for different defects. The last step is to feed predicted bounding boxes into the crack segmenter to produce pixel-level predictions for each road defect. We follow \cite{DBLP:conf/miccai/RonnebergerFB15} to construct the crack segmenter, which is specially trained for segmenting road defects. Figure \ref{fig:seg_result} shows one detection result on the top, and two segmentation pixel-level results (shown in two rows) at the bottom. In the bottom segmentation result, the left image is the detected bounding boxes by the crack detector. The image at the mid is the prediction feature map by the crack segmenter. The right image maps predictions back to the original image for visualisation.

\vspace{-2ex}
\subsection{Road Marking Parsing}
We develop a road marking segmenter to detect different categories of road markings at pixel level accurately. The road marking segmenter follows \cite{DBLP:conf/cvpr/ZhaoSQWJ17} and is thoroughly trained with the existing benchmark dataset \cite{DBLP:journals/pami/HuangWCZGY20} containing various labelled road markings. After prediction, we performed an extra refinement process to fit the results back to the original image. The refinement process can be summarised into three steps. The first step is to detect the region of interest (ROI), which is the road in our case. Then we project both original images and prediction maps from street view into bird eye view to obtain parallel or perpendicular road markings. The second step is to implement low-level contour detection, which focuses on generating the boundaries of objects. The low-level contour of road markings is regular. The predicted pixels are irregular but cover most areas of the road markings. Thus, we compare the prediction map with low-level contours. If there is a high overlap between them, the low-level contours are kept and mapped back to the original image.

\vspace{-5pt}
\section{CONCLUSION}
In this paper, we present a fully developed system - RoadAtlas, that comprehensively provides functions of road defect recognition, road marking parsing integrated into an end-to-end platform. It can achieve faster automated recognition of road defects than manually reviewing images by well-trained inspectors, allowing local governments to repair roads in time. Besides, a great portion of budgets can be saved by full automation. The local governments can effortlessly grasp and manage road conditions with RoadAtlas.

\bibliographystyle{ACM-Reference-Format}
\bibliography{main}










\end{document}